\documentclass[conference]{IEEEtran}
\IEEEoverridecommandlockouts
\usepackage{cite}
\usepackage{amsmath,amssymb,amsfonts}
\usepackage{algorithmic}
\usepackage{graphicx}
\usepackage{textcomp}
\usepackage{xcolor}
\def\BibTeX{{\rm B\kern-.05em{\sc i\kern-.025em b}\kern-.08em
    T\kern-.1667em\lower.7ex\hbox{E}\kern-.125emX}}
    
\ifCLASSOPTIONcompsoc
    \usepackage[caption=false,font=normalsize,labelfon
t=sf,textfont=sf]{subfig}
\else
    \usepackage[caption=false,font=footnotesize]{subfi
g}
\fi

\begin{document}

\title{ TS-RGBD Dataset: a Novel Dataset for Theatre Scenes Description for People with Visual Impairments
}

% \author{
% \IEEEauthorblockN{1\textsuperscript{st} Leyla Benhamida}
% \IEEEauthorblockA{\textit{RIIMA Laboratory, Computer Science Faculty} \\
% \textit{USTHB University}\\
% BP 32 EL ALIA, 16111, Algiers, Algeria \\
% lbenhamida@usthb.dz}
% \and
% \IEEEauthorblockN{2\textsuperscript{st} Khadidja Delloul}
% \IEEEauthorblockA{\textit{RIIMA Laboratory, Computer Science Faculty} \\
% \textit{USTHB University}\\
% BP 32 EL ALIA, 16111, Algiers, Algeria \\
% kdelloul@usthb.dz}
% \and 
% \IEEEauthorblockN{3\textsuperscript{rd} Slimane Larabi}
% \IEEEauthorblockA{\textit{RIIMA Laboratory, Computer Science Faculty} \\
% \textit{USTHB University}\\
% BP 32 EL ALIA, 16111, Algiers, Algeria \\
% slarabi@usthb.dz}
% }

\author{\IEEEauthorblockN{Leyla Benhamida\IEEEauthorrefmark{1},
Khadidja Delloul\IEEEauthorrefmark{2} and Slimane Larabi\IEEEauthorrefmark{3}}
\IEEEauthorblockA{\textit{RIIMA Laboratory, Computer Science Faculty} \\
\textit{USTHB University}\\
BP 32 EL ALIA, 16111, Algiers, Algeria \\
Email: \IEEEauthorrefmark{1}lbenhamida@usthb.dz,
\IEEEauthorrefmark{2}kdelloul@usthb.dz,
\IEEEauthorrefmark{3}slarabi@usthb.dz}}

\maketitle

\begin{abstract}
Computer vision was long a tool used for aiding visually impaired people to move around their environment and avoid obstacles and falls. Solutions are limited to either indoor or outdoor scenes, which limits the kind of places and scenes visually disabled people can be in, including entertainment places such as theatres. Furthermore, most of the proposed computer-vision-based methods rely on RGB benchmarks to train their models resulting in a limited performance due to the absence of the depth modality. 

In this paper, we propose a novel RGB-D dataset containing theatre scenes with ground truth human actions and dense captions annotations for image captioning and human action recognition: TS-RGBD dataset. It includes three types of data: RGB, depth, and skeleton sequences, captured by Microsoft Kinect \footnote{https://github.com/khadidja-delloul/RGB-D-Theatre-Scenes-Dataset}.

We test image captioning models on our dataset as well as some skeleton-based human action recognition models in order to extend the range of environment types where a visually disabled person can be, by detecting human actions and textually describing appearances of regions of interest in theatre scenes.

\end{abstract}


\section{Introduction}
With the advancement known in deep learning technologies, uncountable are applications that emerged in this field. Among these researches, we can find multiple solutions that focus on helping make the life of blind and visually impaired people easier. Either by designing tools to help them move around their environment and detect obstacles and stairs, or by developing applications that help them in their daily life by identifying money bills or objects, reading for them, or offering them online assistance. 

While these applications offer them (blind and visually impaired people) help throughout their daily life transactions and issues, they remain limited when it comes to entertainment. For instance, there are no solutions that help them access and understand a theatre scene by providing a description of the scene and the actors' actions on stage. Even though works that revolve around describing paintings and aesthetics \cite{art1, art2}  or reading books exist,  there are -to our knowledge- no works that are interested in textual descriptions of theatre plays. Although these textual descriptions are sometimes written manually and read by people, they are not always available. 

In this work, we aim to provide blind and visually impaired people with a system that can not only describe a theatre scene for them but to give them the positions of every object or region present on the stage regarding them (left, right, front). To build such a system, we had to use the image captioning 'DenseCap' model to detect regions and generate captions for each one of them, while using depth information to determine their positions regarding the user. However, the first challenge that was encountered was the fact there are no theatre scenes in the images that models are trained on. The second challenge was the absence of depth information from that set of images.

On the other hand, in order to fully comprehend a theatre scene, visually impaired persons need to have a description of the actors' actions performed on stage. This description can be provided after recognizing the actions based on state-of-the-art human action recognition methods. Various techniques have emerged to recognize human actions using a computer vision approach with deep learning models. The emergence of RGB-D sensors, such as the Microsoft Kinect, has revolutionized the field of HAR (human action recognition) by providing rich human action benchmarks\cite{ntu, uwa3d, utkinect} that contain RGB images as well as depth and skeleton information for more accurate action analysis. However, despite significant progress in RGB-D action datasets, there remains a scarcity of datasets specifically designed to capture human actions in theatrical settings. Theatre environments present unique challenges for action recognition due to their distinct characteristics and intricate stage designs.

To address the cited challenges for theatre scene textual description and advance the state-of-the-art in RGB-D human action recognition in a theatre environment, we present a novel dataset specifically tailored for capturing scenes and human actions in theatrical settings containing three modalities: RGB, Depth, and skeleton data. Furthermore, we provide through this dataset two categories of data: trimmed sequences of human actions and untrimmed sequences that represent long continuous theatre scenes with temporal annotation. By introducing our unique dataset with these two categories, we will promote the development of novel techniques capable of not only effectively recognizing actions in theatres but also localizing and detecting the boundaries of actions, using the second category of data, for real-time recognition.

%% Paragraph To verify and rectify
This paper is organized as follows:  Section \ref{sec:reltaedworks} reviews the current benchmarks of both image captioning and human action recognition as well as a small review of the existing approaches for human action recognition and used datasets. Section \ref{sec:dataset} introduces the proposed theatre dataset: TS-RGBD, its structure, annotation process, and detailed information. Then, section \ref{sec:egocap} is devoted to presenting the proposed solution for egocentric captioning, followed by the experimental results of human action recognition models on the proposed dataset, detailed in section \ref{sec:actrecog}. 

\section{Related Works}
\label{sec:reltaedworks}
\subsubsection{Datasets}
Well-known computer vision datasets, even those of considerable acclaim, notably lack theatre images, let alone comprehensive RGB-D data specifically capturing theatre scenes.

The following table gives a summary of available RGB datasets:
    \begin{table}[htbp]
    \begin{center}
    \caption{RGB Datasets.}
    \label{tab:rgb-datasets}
        \begin{tabular}{|c|c|c|}
            \hline
            Dataset Name & Images N° & Type \\
            \hline
            \textbf{MS-COCO} & > 200 000 &  mainly outdoor \\
            \hline
            \textbf{Flickr8K} & 8000 &  outdoor + indoor\\
            \hline
            \textbf{Visual Genome} & 101 174 & outdoor + indoor \\
            \hline
            \hline
            \textbf{Cityscapes} & 25 000 &  urban streets\\
            \hline
            \textbf{ADE20K} & 20 000 &  outdoor + indoor \\
            \hline
        \end{tabular}
    \end{center}
    \end{table}

As for depth datasets:
    \begin{table}[htbp]
    \begin{center}
    \caption{RGB-D Datasets.}
    \label{tab:rgbd-datasets}
        \begin{tabular}{|c|c|c|}
            \hline
            Dataset Name & Images N° & Type \\
            \hline
            \textbf{SUN-RGB-D} & 10 335 & rooms \\
            \hline
            \textbf{NYU-Depth v2} & ~2000 & indoor\\
            \hline
            \textbf{Scan Net} & 1513 & indoor scans \\
            \hline
        \end{tabular}
    \end{center}
    \end{table}
    
From both tables, we conclude that there are no available datasets with theatre plays in them.

\subsection{Image Captioning}
Image captioning consists of describing the content of any given image using text. The automatically generated captions are expected to be grammatically correct, with logical order. Image captioning relies on deep learning models that are based either on retrieval (auto-encoders or features extraction..), template (sentence generation after object detection and recognition), or end-to-end learning \cite{ic2}.

Generated captions can be a single sentence or multiple sentences that constitute a paragraph.   

There are various architectures for single sentence captioning models, from scene description graphs \cite{ic15, ic6} to attention mechanisms \cite{ic16, ic17, ic1, ic4, ic3}, transformers, and even CNN-LSTM and GANs networks  \cite{ic18, ic10, ic14}.

Solutions for paragraph captioning are based on end-to-end dense captioning models. They are based on single-sentence captioning to generate a set of sentences that will be combined to form a coherent paragraph \cite{ic2}. These solutions are built using encoder-decoder architectures and recurrent networks \cite{ic20, ic21, ic3, densecap}. 

Kong et al proposed in \cite{kong} a solution for RGB-D image captioning, but it only focuses on enriching descriptions by positional relationships between objects, while training their model on a dataset that does not include theatre images. 

Whether single sentence or paragraph, image captioning models achieved remarkable results regarding different metrics (BLEU, ROUGE, METEOR, CIDrE..etc). However, they do not generate detailed captions when it comes to complex scenes. Single sentence models focus on moving objects ignoring background, and paragraph captioning models do not consider positional descriptions.

Giving blind and visually impaired people sentences that lack descriptions of static objects and background, or paragraphs that lack positional descriptions of said objects makes it difficult or even impossible for them to re-imagine and rebuild the scene in their minds.

We highlight the fact that most models are trained only on indoor or outdoor scenes, which leads to bad captioning when the images are extracted from theatre scenes.

\subsection{Human Action Recognition}
Human action recognition is a fundamental task in computer vision with numerous applications, ranging from surveillance and human-computer interaction to robotics and virtual reality. Due to its wide range of applications, many methods were proposed that succeeded at achieving considerable performance. The earliest methods were based on RGB sequences \cite{rgb1, rgb2} but their performance is relatively low due to different factors such as illumination and clothing colors. After the release of the Microsoft Kinect sensor, many RGB-D human action benchmarks emerged \cite{ntu, uwa3d, msr} presenting richer information by providing the depth modality resulting in more accurate action features. They mostly consist of three modalities: RGB, depth, and skeleton sequences. As a result, other methods were developed based on the RGB-D datasets that surpass the earliest approaches.  Some methods considered the use of depth maps only \cite{depth1, depth2} which achieved better performance compared to RGB methods but they remain very sensitive to view-point variations. 

Recently, the skeleton-based approach is widely investigated using skeleton sequences and it achieved considerable performance compared to the other approaches, especially after the rise of Graph Convolution Networks (GCN) \cite{sklt1, sklt2, sklt3}. GCNs are designed to extract features from graph-based data such as skeleton sequences that can be modeled as graphs by linking different body joints.
\subsubsection{RGB-D Datasets}
Some of the well-known RGB-D human action benchmarks include:
\begin{itemize}
    \item UWA3D Activity Dataset \cite{uwa3d} contains 30 activities performed at different speeds by 10 people of varying heights in congested settings. This dataset has high inter-class similarity and contains frequent self-occlusions.
    \item MSR Daily Activity3D dataset \cite{msr} includes 16 daily activities in the living room. This dataset can be used to assess the modeling of human-object interactions as well as the robustness of proposed algorithms to pose changes.
    \item MSR Action Pairs \cite{3dpairs} provides 6 pairs of actions in which two actions in a pair involve the interactions with the same object in distinct ways. This dataset can be used to evaluate the algorithms' ability to model the temporal structure of actions.
    \item NTU-RGBD \cite{ntu} was first containing 56880 sequences of 60 action classes. Then, the extended version \cite{ntu120} was introduced with 57367 additional sequences and 60 other action classes making it the largest action benchmark so far. 
\end{itemize}
Most of the proposed benchmarks, including the cited ones, focus only on offline action recognition task that consists of classifying segmented action sequences. However, in the case of real-life applications, temporal localization of actions in untrimmed sequences is very important in order to obtain real-time recognition. In order to elaborate online systems, a few benchmarks were proposed providing a set of untrimmed videos where most of them were collected from Media, TV shows, YouTube...etc, resulting in one modality datasets containing only RGB sequences \cite{tvseries, hac}. Some others were collected using depth sensors, providing multi-modal datasets such as:
\begin{itemize}
    \item G3D \cite{g3d} is intended for real-time action recognition in games with a total of 210 videos. As the first activity detection dataset, the majority of G3D sequences involve several actions in a controlled indoor environment with a fixed camera, which is a typical setup for gesture-based gaming., 
    \item OAD \cite{oad} dataset focuses on both online action detection and prediction. It contains 59 videos of daily actions, and it proposes a set of new protocols for 3D action detection.
    \item PKU-MMD \cite{pku} represents a large-scale dataset containing 1076 sequences with almost 20,000 action instances and 5,4 million frames in total. Besides the three modalities (RGB, Depth, and skeleton sequences), it also provides the corresponding Infrared Radiation data.
\end{itemize}

All of these datasets were captured in either an outdoor environment or an indoor environment (e.g.\ kitchen, room, office...etc), none have considered a theatre environment. 

The task of recognizing human actions in a theatre environment can be very challenging due to its unique characteristics such as dynamic lighting conditions, special stage designs, and complex human interactions.

Therefore, we collect a dataset of RGB-D theatre scenes that contains both trimmed and untrimmed action sequences in order to i) advance the performance of the proposed techniques for both offline and online action recognition in a theatre context, and ii) stimulate the development of novel algorithms and techniques capable of effectively handling the intricacies of theatre environment.

In conclusion, in this work, we make the following contributions: 
\begin{itemize}
    \item To the best of our knowledge, we are the first to collect and provide RGB-D sequences captured in a theatrical setting.
    \item Our dataset provides RGB-D untrimmed theatre scenes with temporal annotations, that contains continuous actors' actions in order to help the development of theatre online action recognition systems.
    \item Image Captions that contain the direction of each region, with captioning model retrained on our theatre scenes dataset.

\end{itemize}

\section{TS-RGBD Dataset Description}
\label{sec:dataset}
In this section, we describe the data collection process, and dataset statistics in detail as well as annotation and cleaning methodologies.

\subsection{Setup}
In order to collect samples in a theatre environment, we sought cooperation with national theaters. Thus, we contacted the UK National Theater, but because of the terms of the actors' contracts, it was not possible to use their visual content. Our local National Theater on the other hand was open for a partnership with the laboratory to achieve the task. However, the limited range of the Kinect sensor hindered us from accurately capturing the depth information of actors situated at a distance beyond four meters. 
 
\begin{figure}[htpb]
        \centerline{\includegraphics[width=2in]{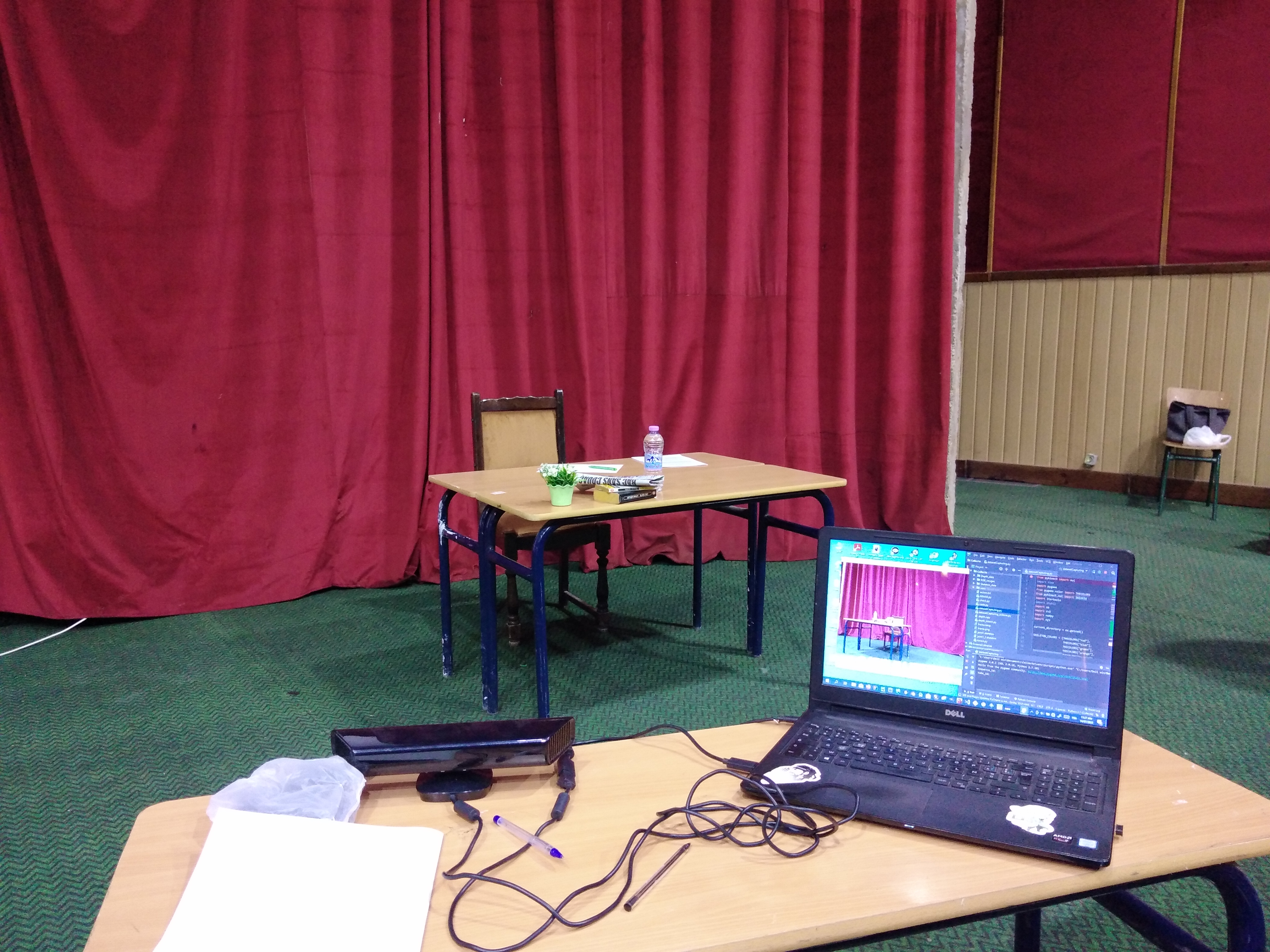}}
        \caption{Scene capturing in the university auditorium.}
        \label{fig:auditorium}
    \end{figure}

Finally, we opted to film various scenarios at the auditorium of the university (figure \ref{fig:auditorium}) where the distances are convenient for the Kinect sensor.

\begin{figure}[htpb]
        \centerline{\includegraphics[width=2in]{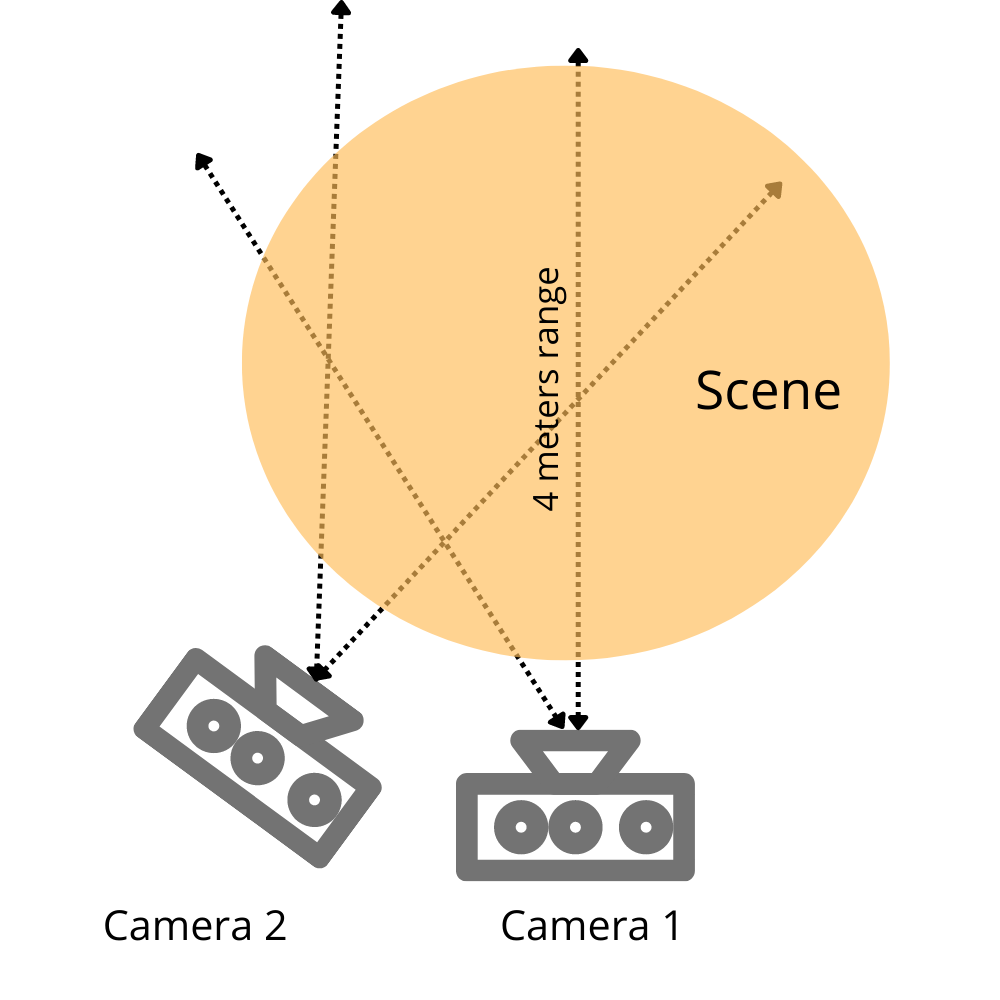}}
        \caption{Illustration of the Kinects setup.}
        \label{fig:setup}
    \end{figure}

Two Kinect v1 sensors were used and positioned at the same height in two different viewpoints (front view and side view) as shown in  Figure \ref{fig:setup}. We also used more than 76 objects in total to vary the setups and the used/background objects. The use of two sensors at different positions and varying background setups results in the diversity of the collected samples.

\subsection{Subjects}
We enlisted a team of 8 students to interpret on stage the prepared scenarios. The students signed a legal document granting us permission to use and distribute their visual content among the scientific society.
 
\subsection{Data Modalities}
The Microsoft Kinect v1 provides three data modalities: RGB images, depth, and skeleton data.

The resolution of each captured RGB and depth sequence is $640\times480$, and each frame is saved in JPEG format. The sequences were captured at a rate of 25 frames per second.

\begin{figure}[ht]
    \centering
    \includegraphics{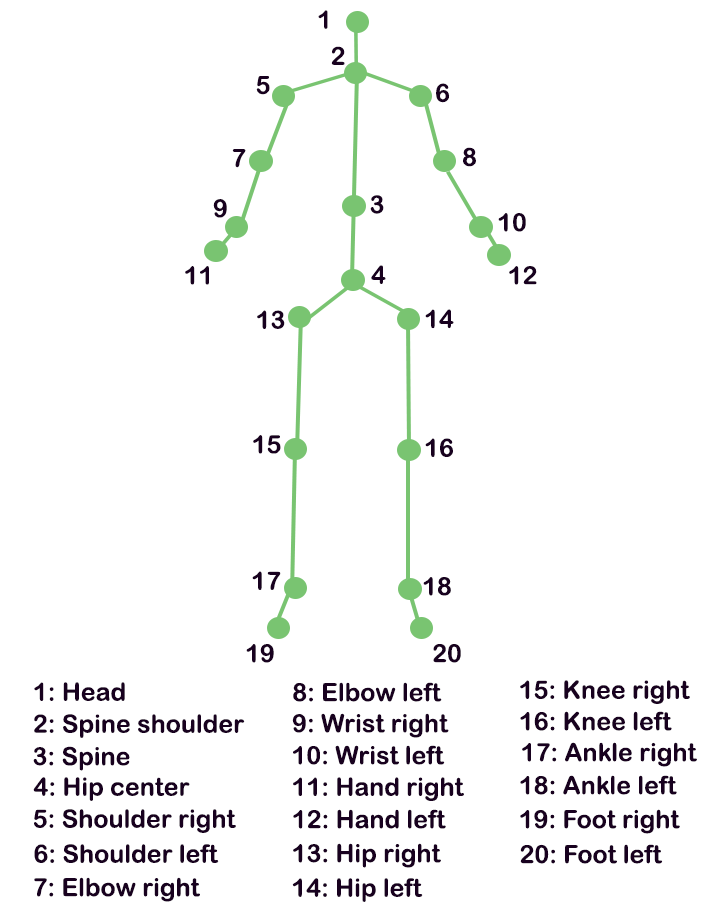}
    \caption{Joints configuration provided by Kinect v1.}
    \label{fig:joints}
\end{figure}

The skeleton data, on the other hand, consists of 3-dimensional positions of 20 body joints for each tracked human body, knowing that Kinect v1 can only detect and track at most two human bodies. Figure.\ref{fig:joints} illustrates the configuration of the 20 captured joints.

\subsection{Data Classes}
Our dataset consists of two categories of data: \textit{segmented theatre actions} and \textit{untrimmed theatre scenes}.

\subsubsection{Segmented theatre actions}
This category contains 36 action classes that are more accurate in theatre scenes such as walking, sitting down, drinking, jumping, eating, and throwing. Each viewpoint comprises 230 sequences, with an average of 170 frames for each sequence. Each action was carried out by 3 males and was repeated at least 3 times at varying speeds.
%add figures

\subsubsection{Untrimmed theatre scenes}
This category includes 38 written theatre scene scenarios. It contains, in total, 75 sequences for each viewpoint, with a mean of 1119 frames per sequence. The scenes are  divided into three types:
\begin{itemize}
    \item \textbf{Solo scenes} involve a single person performing different actions (figure \ref{fig:scenarios1}). Each solo scene was interpreted by at least two individuals to ensure data diversity.
    \begin{figure}[ht]
        \centerline{\includegraphics[width=3in]{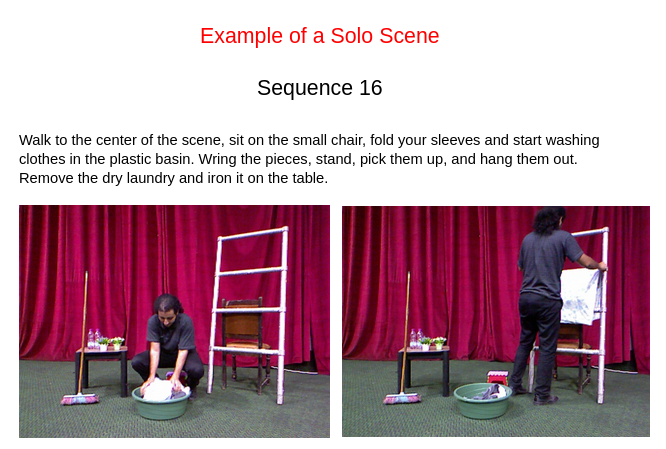}}
        \caption{Example of interpreted scenarios (Solo).}
        \label{fig:scenarios1}
    \end{figure} 

    \item \textbf{Two-Person Scenes} involve interactions between two individuals, such as "two persons walking towards each other", "shaking hands", "one person handing an object to another one", and "hugging each other" as shown in figure \ref{fig:scenarios2}.
    \begin{figure}[ht]
        \centerline{\includegraphics[width=3in]{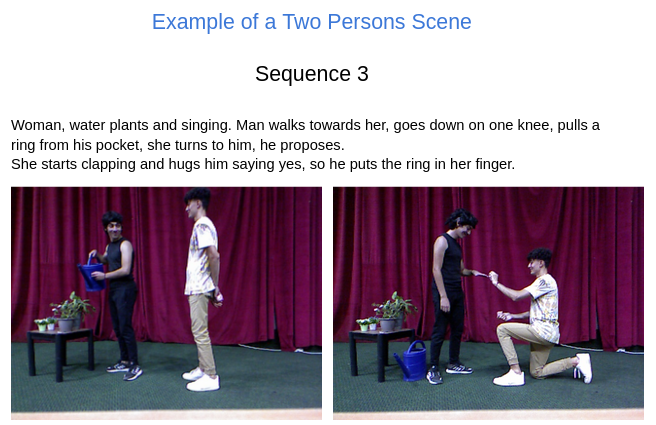}}
        \caption{Example of interpreted scenarios (Two People).}
        \label{fig:scenarios2}
    \end{figure}

    \item \textbf{Group Scenes} involve three or more people engaged in an activity. Notably, skeleton data of this last type of scene is considered as a two-person interaction scene because, as mentioned before, Kinect v1 can only track skeleton joints of at most two persons. Figure \ref{fig:scenarios3} shows an example of such scenes.
    \begin{figure}[ht]
        \centerline{\includegraphics[width=3in]{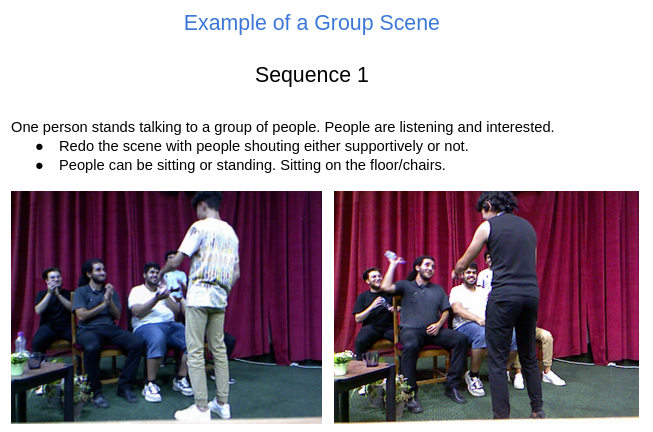}}
        \caption{Example of interpreted scenarios (Group).}
        \label{fig:scenarios3}
    \end{figure}
\end{itemize}

Summarily, with 8 male actors (females were not available) we could gather 610 sequences with an average of 373 frames per sequence (25 frames per second), and a total of 123 149  frames.

% Depth data was stored in a JPEG file for each frame. We used 76 objects as well. ( This is already mentionned in "Setup" section)

The table \ref{tab:stats} presents a summary:
    
    \begin{table}[htbp]
    \begin{center}
    \caption{Captured RGB-D data from written scenarios. }
    \label{tab:stats}
        \begin{tabular}{|c|c|c|}
            \hline
            Sequences Tot. & Frames Tot. & N° used Objects\\
            \hline
            610 & 123 149   & 76 \\
            \hline
        \end{tabular}
    \end{center}
    \end{table}

\begin{figure}[htpb]
        \centerline{\includegraphics[width=3in]{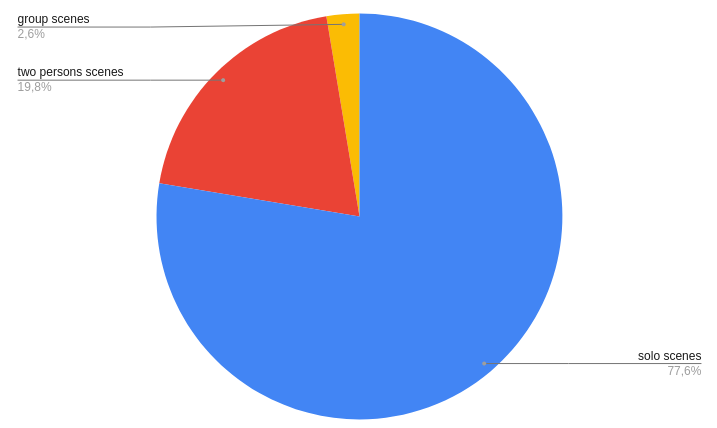}}
        \caption{Pie chart for the number of sequences by type of scenario.}
        \label{fig:stats}
    \end{figure}

Figure \ref{fig:stats} shows the number of sequences per type of scenario. There are more solo scenes since the Kinect v1 range is limited to $4$ meters and resolution (640$\times$480) which makes it impossible to fit a group of people into such a small frame due to their height differences.

\subsection{Data Cleaning}
For the image captioning task, we created an application to manually select frames with smooth depth maps, that mark a transition in the video to avoid redundancies.

In addition to that, we had to go over all selected frames to keep only the ones with smooth corresponding depth maps.

In the end, 1480 key-frames were kept.

\subsection{Data Annotation}

Many data annotation applications available today offer powerful functionality for annotating data, but they often come with a trade-off: either our data become publicly accessible, or these applications come at a cost and are not available for free.

Even so, we could find a multi-platform desktop application developed by \cite{labelme} available to download and install from GitHub. The developer was inspired by the original \textit{"LabelMe"} application that was created by MIT for manually annotating data for object detection/recognition and instance or semantic segmentation, with the possibility of drawing a box or a polygonal envelope and adding labels.

We could annotate 50 images so far, resulting in the following:
    \begin{table}[htbp]
    \begin{center}
    \caption{Numbers of Annotated Data.}
    \label{tab:annotate}
        \begin{tabular}{|c|c|c|}
            \hline
            Regions N° & Captions N° & Tokens N° \\
            \hline
            504 & 504  & 109 \\
            \hline
        \end{tabular}
    \end{center}
    \end{table}

Figure \ref{fig:labelme} shows the interface of the \textit{"LabelMe"} application as well as the process of polygonal annotations:

    \begin{figure}[htpb]
        \centerline{\includegraphics[width=3in]{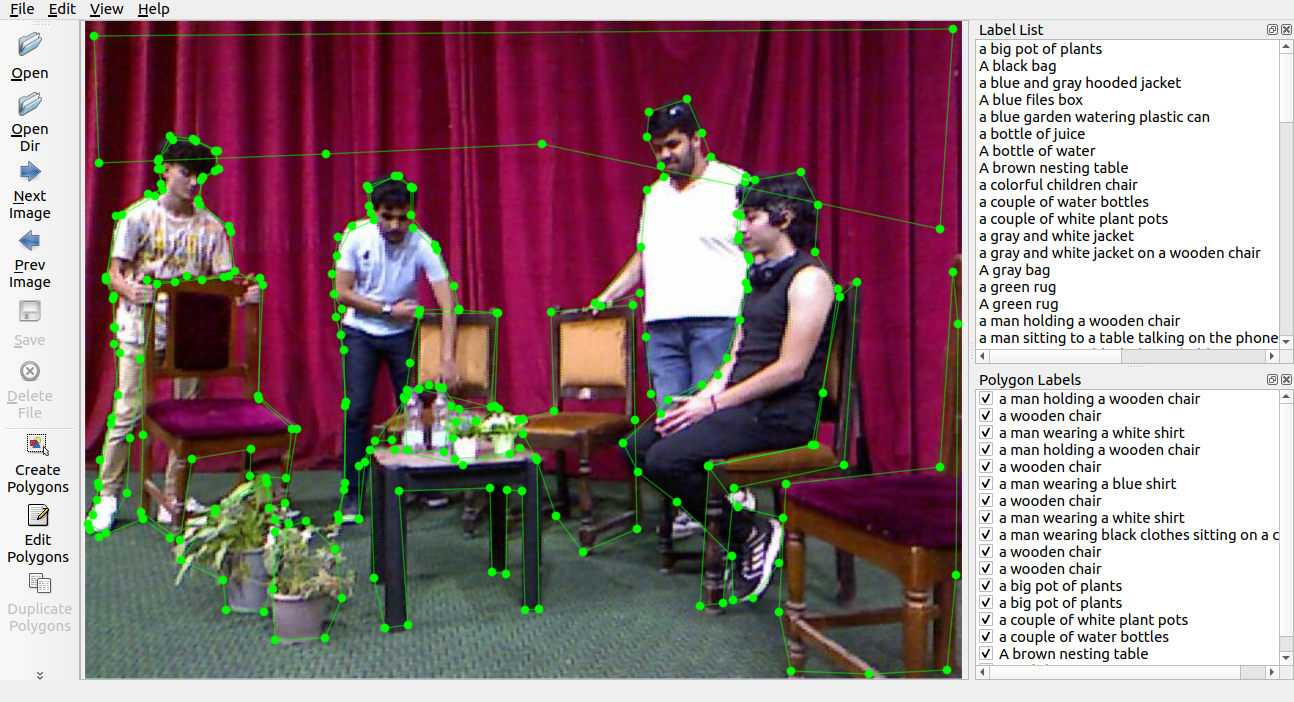}}
        \caption{LabelMe Interface.}
        \label{fig:labelme}
    \end{figure}

\section{Egocentric Captioning}
\label{sec:egocap}
\subsection{Proposed Solution}
In this paper, we propose an approach to offer the blind
and visually impaired detailed descriptions of the environment
they are in while giving them the opportunity to attend theatre plays.
Those descriptions will be generated by the DenseCap module
that outputs captions for both mobile and static objects and
regions in a given scene. These generated captions are not
enough for the users to re-imagine the scene, they will need
to know where each object or region is situated regarding
their own position (Egocentric Description). To give the users
this information, we will need depth data alongside RGB
image of the scenes, specifically theatre scenes. 

\begin{figure}[htpb]
        \centerline{\includegraphics[width=2.5in]{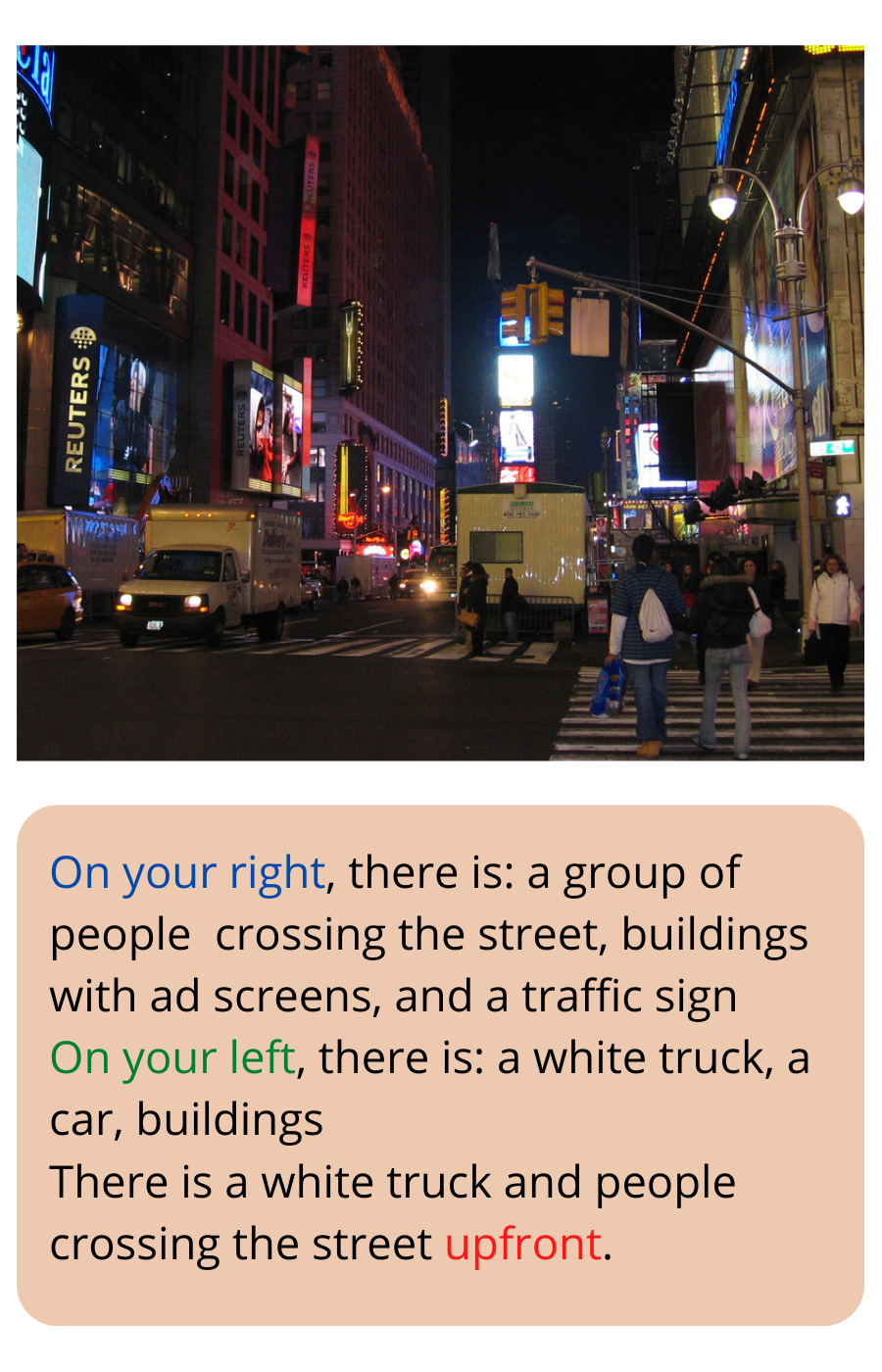}}
        \caption{Example of Egocentric Scene Description.}
        \label{fig:expect}
    \end{figure}

An example of the expected description is shown in figure \ref{fig:expect}.
    
To do that, we had to retrain the DenseCap model on our dataset. Proposed in \cite{densecap}, it is a model based on Fully Convolutional Localization Networks (FCLN) that outputs boxes surrounding detected regions, each box with its caption and confidence. We chose DenseCap because it does not focus only on salient objects and provides background descriptions. 

After detecting regions and generating the corresponding captions, we applied the algorithm proposed in our precedent work to get the directions \cite{mine1}.

Since Depth information is not present for the VG dataset, we used AdaBins model to estimate depth maps for VG images.

\subsection{Experiments and Results}
We modified the DenseCap code provided in GitHub to be trained on custom data and we applied transfer learning by reusing the models' weights provided by the authors to train it on our data for 10 more epochs. 

Table \ref{tab:eval} shows evaluation results after using DenseCap on our data before and after retraining.

    \begin{table}[htpb]
    \begin{center}
    \caption{Captions Evaluation.}
    \label{tab:eval}
        \begin{tabular}{|c|c|c|c|c|}
            \hline
            Case & METEOR & BLEU & ROUGE & CIDEr \\
            \hline
            Before Re-Training & 0.21 & 0.30 & 0.32 & 1.25 \\
            \hline
            After Re-Training & 0.52 & 0.6 & 0.63 & 5.52 \\
            \hline
        
        \end{tabular}
    \end{center}
    \end{table}

We then chose 20 random images from VG and our dataset to manually annotate the direction of each generated region.

 The table \ref{tab:dirAccOurs} summarizes results.

    \begin{table}[htpb]
    \begin{center}
        \caption{Egocentric Description Evaluation on Our Images.}
        \label{tab:dirAccOurs}
        \begin{tabular}{|c|c|c|c|}
            \hline
            Dataset & Correct Directions & Incorrect Directions & Acc.  \\
            \hline
            \textbf{Ours} & 195 & 5 & \textbf{97.5\%} \\
            \hline
            VG & 175 & 21 & 89\% \\
            \hline
        \end{tabular}
    \end{center}
    \end{table}

\begin{figure*}[!t]
\centering
\subfloat{
    \subfloat{\includegraphics[height=1in,width=1.5in]{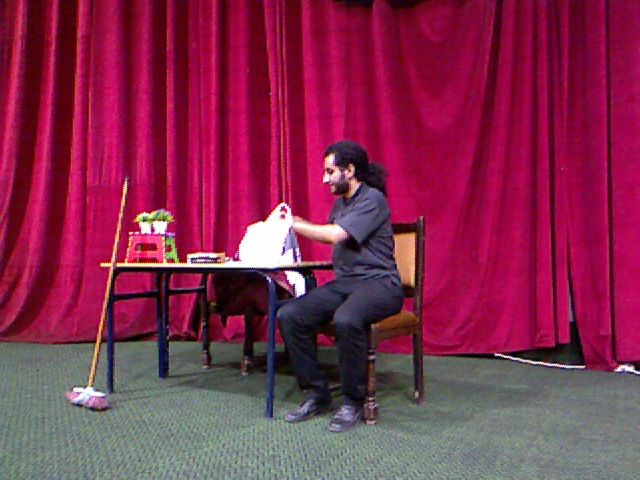}}
    \subfloat{\includegraphics[height=1in,width=1.5in]{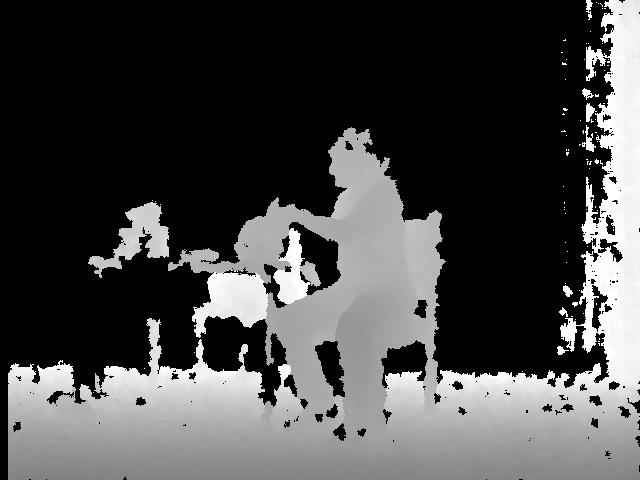}}
    \subfloat{\includegraphics[height=1in,width=1.5in]{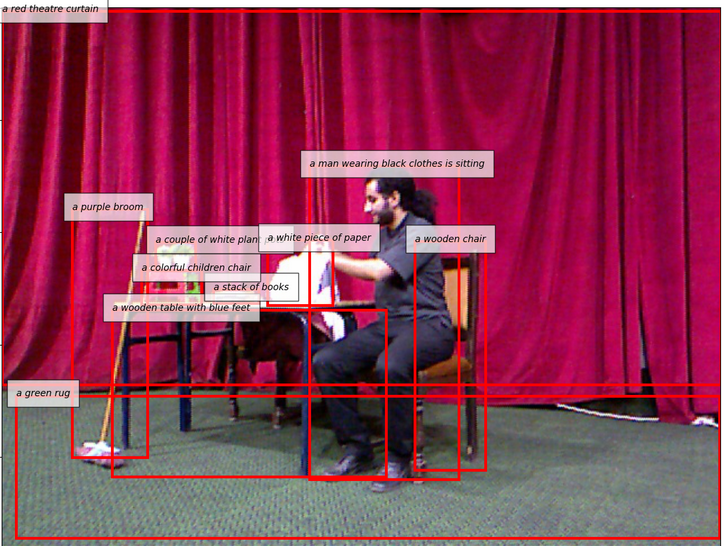}}
    \subfloat{\includegraphics[height=1in,width=1.5in]{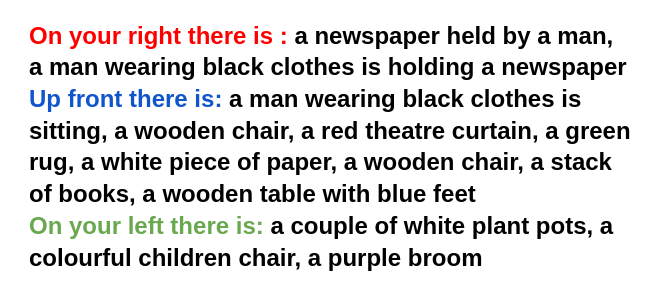}}
 }
\hfil
\subfloat{
    \subfloat{\includegraphics[height=1in,width=1.5in]{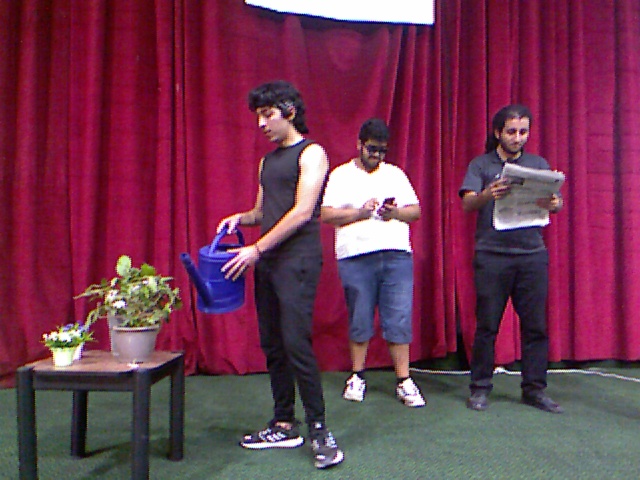}}
    \subfloat{\includegraphics[height=1in,width=1.5in]{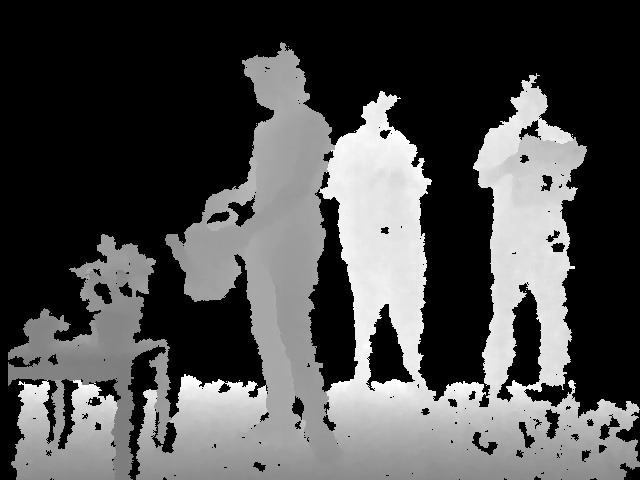}}
    \subfloat{\includegraphics[height=1in,width=1.5in]{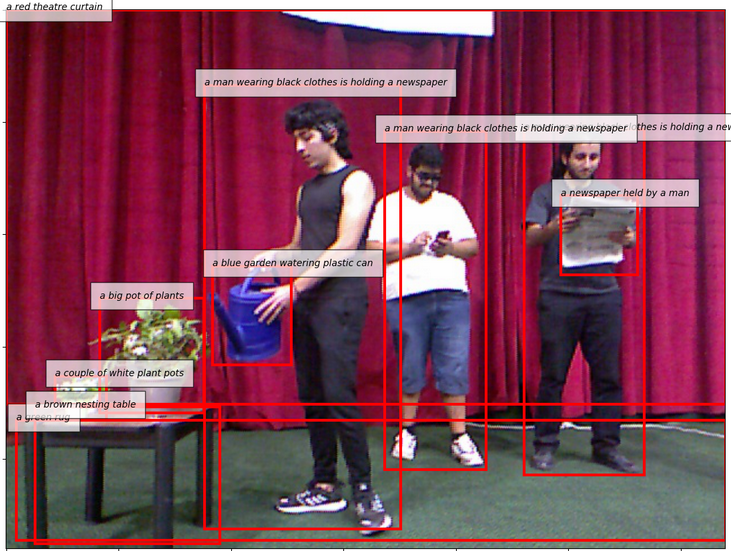}}
    \subfloat{\includegraphics[height=1in,width=1.5in]{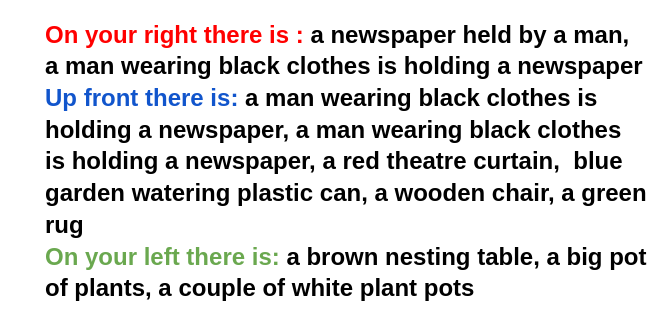}}
    
}
\hfil
\subfloat{
    \subfloat{\includegraphics[height=1in,width=1.5in]{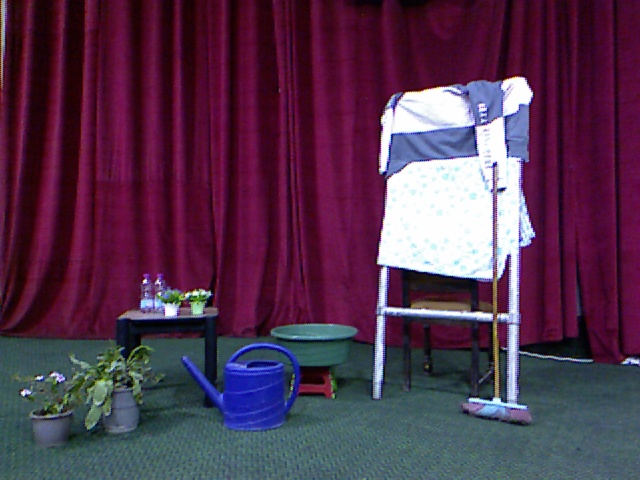}}
    \subfloat{\includegraphics[height=1in,width=1.5in]{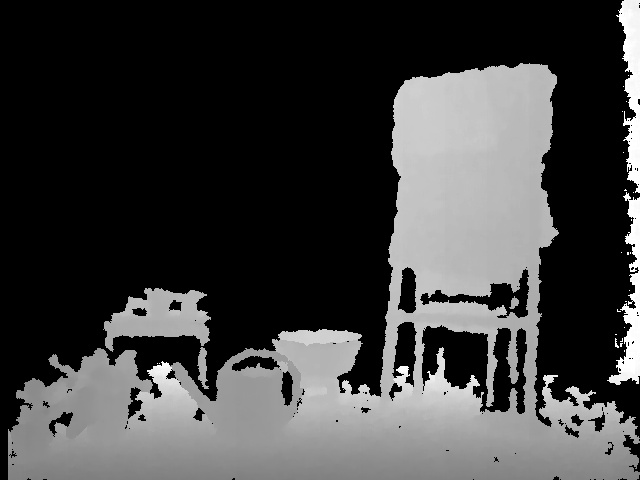}}
    \subfloat{\includegraphics[height=1in,width=1.5in]{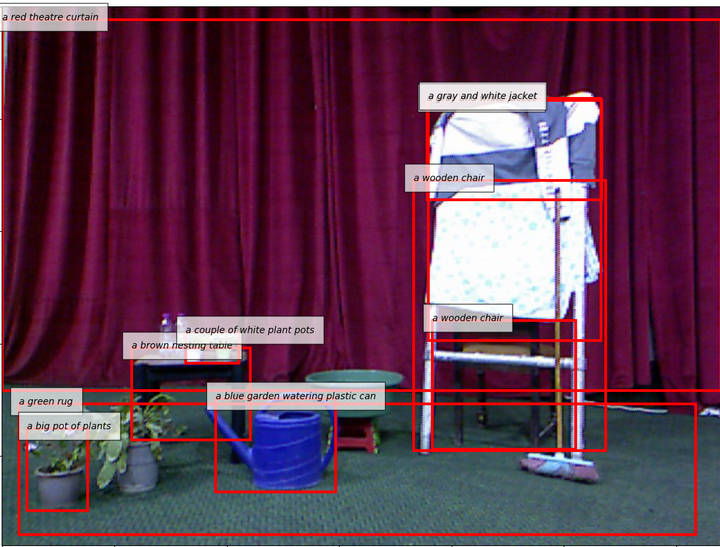}}
    \subfloat{\includegraphics[height=1in,width=1.5in]{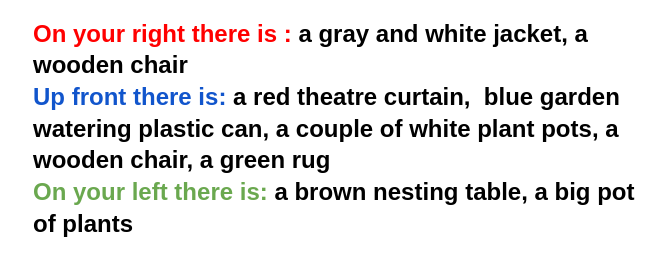}}
}
\hfil
\subfloat{
    \subfloat{\includegraphics[height=1in,width=1.5in]{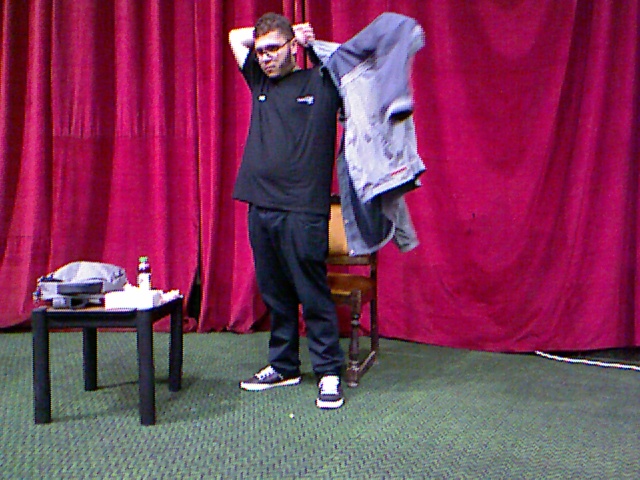}}
    \subfloat{\includegraphics[height=1in,width=1.5in]{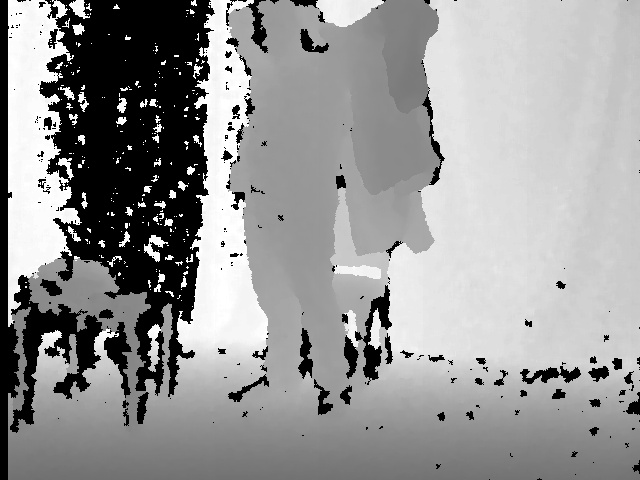}}
    \subfloat{\includegraphics[height=1in,width=1.5in]{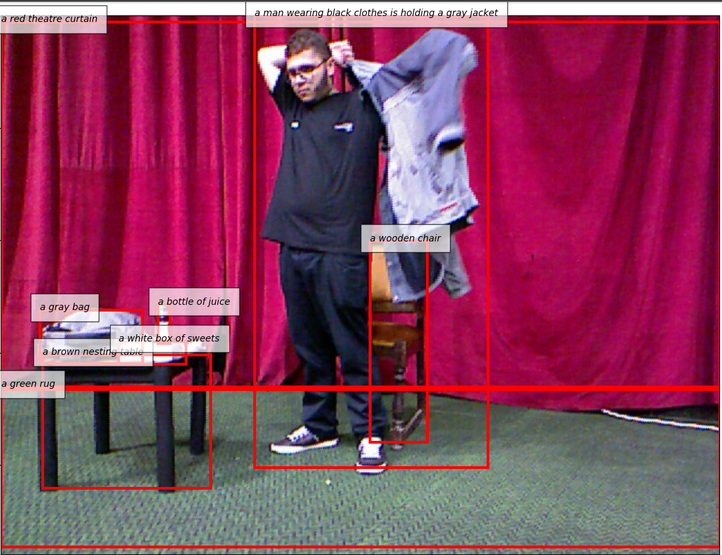}}
    \subfloat{\includegraphics[height=1in,width=1.5in]{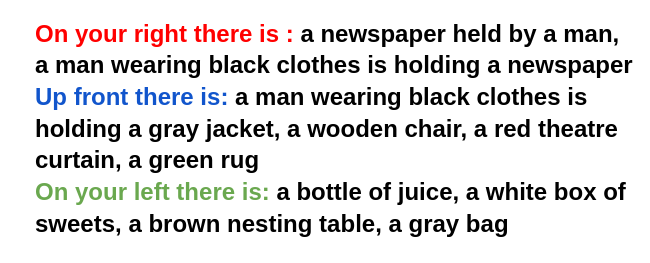}}
}
\hfil
\caption{Multiple Examples from TS-RGBD dataset.}
\label{fig:results}
\end{figure*}

Qualitative results are shown in the figures \ref{fig:results}.

\subsection{Limitations}
\begin{itemize}
    \item Captions are redundant due to the fact that DenseCap generates $k$ number of captions and $k$ was set to 10. Sometimes there are fewer regions than 10, and sometimes there are more, which cannot be possible to determine by a visually impaired person.
    \item Egocentric description lacks precision for some regions.
    \item Final description doesn't mention that the image is about a theatre play. 
\end{itemize}

\section{Human Action Recognition: Experimental Evaluations with TS-RGB}
\label{sec:actrecog}
We conducted experiments using the proposed theatre dataset using its skeleton sequences (Fig. illustrates some of the skeleton sequences of TS-RGBD) on skeleton-based approach with three Graph Neural Networks: ST-GCN\cite{sklt1},  2S-AGCN\cite{2s-agcn}, and MS-G3D\cite{ms-g3d}. We selected to test skeleton-based GCNs due to their highly attained performance. All of the selected models fall under spatio-temporal models which can extract both spatial and temporal features from skeletal sequences. They were mostly trained on NTU-RGBD\cite{ntu}  and Kinetics\cite{kinetics} and the relevant results are illustrated in Table.\ref{tab:acc_ntuKinetics}, which demonstrate their high recognition performances on these very challenging benchmarks.
\begin{figure}[h]
    \centering
    \includegraphics[scale=0.13]{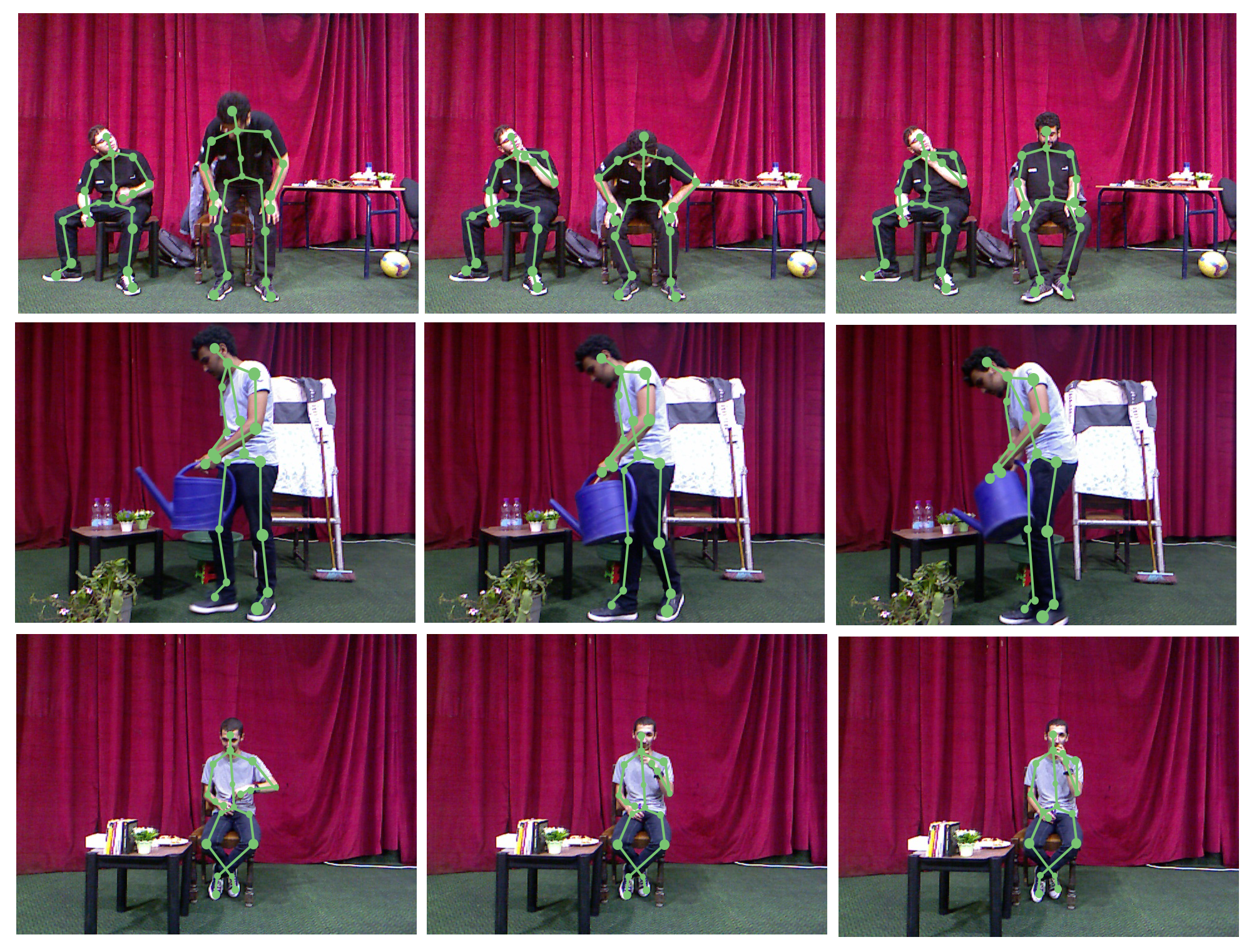}
    \caption{Examples of Skeleton data sequences from TS-RGBD dataset.}
    \label{fig:sklt_data}
\end{figure}
\begin{table}[ht]
    \begin{center}
        \caption{Obtained accuracies by ST-GCN, 2s-AGCN, and MS-G3D on NTU-RGBD and Kinetics.}
        \label{tab:acc_ntuKinetics}
        \begin{tabular}{|c|c|c|}
            \hline
                & NTU-RGBD & Kinetics.  \\
            \hline
            ST-GCN\cite{sklt1} & 81.5\%  & 30.7\% \\
            \hline
            2s-AGCN\cite{2s-agcn} & 88.5\% & 36.1\% \\
            \hline
            MS-G3D\cite{ms-g3d} & 91.5\% & 38.0\% \\
            \hline
        \end{tabular}
    \end{center}
    \end{table}
Thus, we use the available pre-trained weights ( after the training on NTU-RGBD dataset) of each model and test it on our dataset. We obtained the results shown in Table.\ref{tab:acc_ts}.
\begin{table}[ht]
    \begin{center}
        \caption{Test results of ST-GCN, 2s-AGCN, and MS-G3D with TS-RGBD.}
        \label{tab:acc_ts}
        \begin{tabular}{|c|c|}
            \hline
                & Accuracy.  \\
            \hline
            ST-GCN\cite{sklt1} & 50.01\% \\
            \hline
            2s-AGCN\cite{2s-agcn} & 55.73\%  \\
            \hline
            MS-G3D\cite{ms-g3d} & \textbf{60.96\%} \\
            \hline
        \end{tabular}
    \end{center}
    \end{table}

\subsection{Discussion}
\begin{figure}[ht]
    \centering
        \includegraphics[scale=0.7]{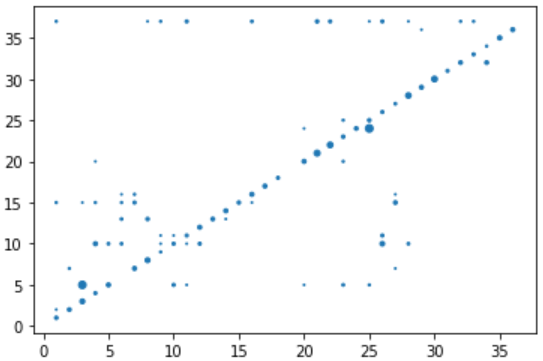}
        \caption{ Confusion Matrix of MS-G3D with TS-RGBD. $y=37$ represents actions of NTU-RGBD that are not included in our dataset}
        \label{fig:cfm}
    \end{figure}

We observe that the performances of the models on our dataset are relatively low. MS-G3D outperformed the other models, so we pursued more comprehensive data from its experiment by analyzing its confusion matrix and extracting the most well-classified as well as misclassified action classes (Table.\ref{tab:wc_mc} and Figure.\ref{fig:cfm}).

    \begin{table}[ht]
    \begin{center}
        \caption{Most well-classified and misclassified action classes.}
        \label{tab:wc_mc}
        \begin{tabular}{|c|c|}
            \hline
               Top 10 well-classified classes & 
               Top 10 misclassified classes.  \\
            \hline
            \hline
            Put on shoes & Drop\\
            Put on jacket &	Put palms together \\
            Walk &	High five with a person \\
            Punch/Slap &	Falling down \\
            Hug a person &	Drink \\
            Walking apart from each other & Clap \\ 
            Walking towards each other	& Write \\
            Sit-down &	Stand-up \\
            Take off jacket &	Check-time  \\
            Push a person &	Read \\

            \hline
        \end{tabular}
    \end{center}
    \end{table}

Based on Table.\ref{tab:wc_mc} and Figure.\ref{fig:cfm}, we distinguish that the model is somewhat weak in recognizing actions that require details about specific body parts, such as the hand shape, or about the involved object in the case of human-object interaction. For instance, the action "write" necessitates additional information on the hand form and the used object, which are not included in the skeleton representation. As a result, it was frequently confused with the action "play with phone" due to the similarity of their skeleton motion trajectories. The same is true for the action 'Drop,' which the model failed to recognize due to missing information about the dropped object and similarities in skeleton motion with other actions, making it difficult to differentiate them based solely on skeleton joint positions.

In conclusion,  there are two major elements that have a large impact on the skeleton-based approach recognition performance. The first factor is the precision of the provided joints' positions. The recognition performance can be low if the skeleton joints are not very well captured and cluttered. The second factor is the number of characteristics that can be extracted from the skeleton modality only. It is not sufficient to recognize some actions that require details about specific body parts' characteristics such as hands or about the involved object in the case of human-object interaction.\\

Future work on our dataset may consider combining skeleton modality with other modalities as a solution to the lack of information problem, which may aid in differentiating between some confusing actions with similar skeleton motions.

\section{Conclusion}
In conclusion, this paper presents the TS-RGBD dataset, a novel RGB-D dataset containing theatre scenes with ground truth human actions and dense captions annotations. The dataset includes RGB, depth, and skeleton sequences captured using the Microsoft Kinect sensor. The purpose of this dataset is to help address the limitations of existing computer vision solutions for aiding visually impaired individuals, which are often limited to either indoor or outdoor scenes, excluding certain environments like theatres.

By incorporating depth information along with RGB data, the TS-RGBD dataset aims to improve the performance of image captioning and human action recognition models. The inclusion of depth modality allows for a more comprehensive understanding of the scenes and actions, enhancing the capabilities of computer vision models to describe the appearances of regions of interest and recognize human actions accurately.

The results of testing image captioning models and skeleton-based human action recognition models on the TS-RGBD dataset demonstrate its potential to expand the range of environment types where visually disabled individuals can navigate with the aid of computer vision technology. The combination of accurate human action recognition and textual description of theatre scenes can provide valuable assistance to visually impaired individuals in accessing entertainment places and enjoying theatrical experiences.

In summary, the TS-RGBD dataset and the discussed methods in this paper contribute to the advancement of computer vision applications for assisting visually impaired individuals, particularly in theatre settings. The dataset's availability and the performance of the tested models open up new possibilities for developing more inclusive and versatile assistive technologies, making entertainment venues and various other environments more accessible to visually disabled individuals. However, further research and development are required to optimize and generalize these methods for real-world applications and potentially adapt them to other challenging scenarios beyond theatre scenes.

\end{document}